# Credit Card Fraud Detection Using Enhanced Random Forest Classifier for Imbalanced Data

AlsharifHasan Mohamad Aburbeian, and Huthaifa I. Ashqar*
Department of Natural, Engineering and Technology Sciences, Arab American University
Ramallah P600, Palestine
a.aburbeian@student.aaup.edu, huthaifa.ashqar@aaup.edu

**Abstract**

The credit card has become the most popular payment method for both online and offline transactions. The necessity to create a fraud detection algorithm to precisely identify and stop fraudulent activity arises as a result of both the development of technology and the rise in fraud cases. This paper implements the random forest (RF) algorithm to solve the issue in the hand. A dataset of credit card transactions was used in this study. The main problem when dealing with credit card fraud detection is the imbalanced dataset in which most of the transaction are non-fraud ones. To overcome the problem of the imbalanced dataset, the synthetic minority over-sampling technique (SMOTE) was used. Implementing the hyperparameters technique to enhance the performance of the random forest classifier. The results showed that the RF classifier gained an accuracy of 98% and about 98% of F1-score value, which is promising. We also believe that our model is relatively easy to apply and can overcome the issue of imbalanced data for fraud detection applications.

## I. Introduction

Financial fraud is a serious problem that is only getting worse and has far-reaching effects on the financial sector, businesses, and the government [1]. Fraud is defined as criminal deception done with the intention of making money [2]. Credit card transactions have surged thanks to a high reliance on internet technology. The rate of credit card fraud is rising as credit card transactions take over as the preferred method of payment for both online and offline transactions [3]. There are two types of credit card fraud: internal and external [4]. While external card fraud entails using a stolen credit card to obtain money through illegal ways, inner card fraud happens as a result of an agreement between cardholders and the bank and involves using a fake identity to commit fraud. Most credit card frauds are external card fraud, which has been the subject of much investigation. Another classification has been made into three categories: classic card-related frauds (application, stolen, account takeover, fake, and counterfeit), frauds involving retailers (merchant collusion and triangulation), and frauds involving the internet (site cloning, credit card generators, and false merchant sites) [5].

Due to their time-consuming nature and ineffectiveness, manual methods of fraud detection have become increasingly impracticable with the introduction of big data. The challenge of credit card fraud, however, has drawn the attention of financial institutions to current computational approaches. One significant way for detecting credit fraud is the use of data mining techniques. The technique of separating fraudulent transactions into two categories: legitimate and fraudulent transactions is known as credit card fraud detection [1]. Machine learning has many techniques to solve this problem including but not limited to Artificial Neural Networks (ANNs), Naive Bayesian Classifier, support vector machines (SVMs), decision trees, random forests, k-nearest neighbors, logistic regression, intelligent decision engines (IDE), meta-learning strategy [6]–[12].

The remainder of the page is divided as follows: discussing the related studies and addressing the research gap in Section II. Dataset acquisition, preprocessing, and feature selection are discussed in Section III. The methodology is illustrated in Section IV. Showing the results and metrics used to improve the reliability of the proposed classifier in section V. Finally, we provided conclusion in Section VI.

## II. Literature review

### A. Related work

Credit card transaction classification is typically a binary classification issue [13]. Here, a credit card transaction falls into one of two categories: either it was fraudulent (negative class) or it was legal (positive class). Data from credit card transactions are primarily distinguished by an unusual phenomenon. Transactions that are both legitimate and fraudulent frequently have the same characteristics. Fraudsters continually find new ways to imitate the spending habits of legitimate cards. The characteristics of legitimate and fraudulent behavior are thus continually changing. Due to this innate property, there are fewer actual fraudulent cases detected in a pool of credit card transaction data, which leads to a distribution that is significantly skewed in favor of the negative class (legitimate transactions). For example, the dataset examined by [16], [17], and [9] contains 20%, 0.025%, and 0.172% respectively fraud cases in all transactions. The highly skewed credit card transaction data have been sampled using a variety of methods. Using a random sample methodology such as [16], [18] presents experimental findings showing that classifiers with the highest true positive rate and the lowest false positive rate are produced by artificially distributing training data in a 50:50 ratio between fraud and non-fraud. For a meaningful number, the study of [9] employs stratified sampling to under-sample the valid records. With reference to performance comparisons on the 1:99 set, the experiment on 50:50, 10:90, and 1:99 distributions of fraud to valid instances finds that the 10:90 distribution performs better since it is most representative of the actual distribution of frauds and legitimate cases. Stratified sampling is used in [19] as well, in order to preserve important patterns from the data, a combination of under-sampling the negative cases and oversampling the positive cases is used in this study. The goal of fraud detection is typically seen as a data mining classification challenge, where the classification of credit card transactions as legal or fraudulent must be done correctly [14], [15].

Typically, there are two different ways for credit card fraud detection in the real world. Firstly, using data mining to manually detect fraud, which is an inaccurate and time-consuming process. The second one is the use of rule-based systems also known as machine learning techniques or Expert systems which are used to store and modify fraud knowledge in order to interpret the information in a meaningful way to combat fraud events. These expert systems can be classified into supervised, unsupervised, and semi-supervised [16]. In order to categorize new transactions as fraudulent or legitimate in supervised fraud detection methods, models are estimated based on samples of fraudulent and legitimate transactions, whereas in unsupervised fraud detection, outliers' transactions are identified as potential examples of fraudulent transactions [17]. While combining both supervised and unsupervised is called a semi-supervised technique. Many studies implement the random forest algorithm to solve the issue in the hand. For example, a brief study conducted by [18], [19] according to these studies, their classifier has an accuracy of 90%, and 98.6% respectively for detecting fraud and non-fraud transactions. In the study of [20], the behavior characteristics of normal and fraud transactions are trained using two different types of random forests algorithms which differ in their underlying classifiers, the best algorithm gains an accuracy of 96.77%, precision, recall, and F1 score of 89.46%, 95.27%, and 96.1% respectively. In the study [21], the researchers compare many algorithms for fraud detection. The results show that the random forest classifier accuracy was 99.7%, precision, recall, and F1 score for class 1 which indicates the fraud transactions were 31%, 89%, and 46% respectively.

Another comparative study [22] investigates different classification algorithms for highly skewed dataset namely logistic regression, random forest, decision trees, and naïve Bayes. According to the results, the random forest classifier has the best performance with an accuracy of 96.77%, precision of 100%, recall of 91.11%, and F1 score of 95.43%. Comparing the performance of random forest, logistic regression, and support vector machine was conducted by [23]. In this study, the researchers use the area under the ROC curve (AUC) and average precision (AP) as metrics to evaluate the mentioned algorithms. According to the results, the random forest classifier gave the best performance with an AUC score of 91.48%, and an AP of 84.83%. Another comparative study [24], investigates the efficiency of logistic regression, k-nearest neighbor, random forest, support vector machine, and decision tree. After implementing the five models the best one was the random forest model with an accuracy of 88%. The main problem when dealing with credit card fraud detection is the imbalanced dataset in which the majority of the transaction are non-fraud ones. This issue makes any supervised learning algorithms able to predict only non-fraud transaction transactions. Although the importance of the F1 score is considered the most real metric to evaluate algorithm performance [25]–[27].

### B. Research gap

The main problem when dealing with credit card fraud detection is the imbalanced dataset in which the majority of the transaction are non-fraud ones. Mentioned studies tried to solve this problem using a supervised learning algorithm but they showed only accuracy which is not sufficient to address the credit card issue. Because the implemented classifier will be obviously biased to the major class and unable to predict the fraud transaction. The classifier's performance should be evaluated in terms of other metrics like precision, recall, and F1-score, to get a better understanding of the classifier's performance. Therefore, this study aims to implement an enhanced binary random forest classifier with the ability to predict both classes (fraud and non-fraud) accurately and measure the result using different metrics to improve the classifier's performance.

### III. Dataset

The dataset used in this study contains details of about 94,682 online credit card transactions labeled as fraud or legitimate ones. After collection, the dataset was masked according to the customer's privacy issues and published on the Kaggle website [28]. It contains 20 features shown in Table 1, which illustrates a sample of the dataset. The domain feature is the domain name of the customer's email address that was used for the transaction (masked), the state feature is the state code of the customer's location, the ZIP code feature is the zip code of the customer's location, time1 and time 2 features illustrating the beginning and the finish time of the transaction, TRN_AMT feature is the transaction amount, TOTAL_TRN_AMT is the total transaction amount, TRN_TYPE is the classification of the transaction (fraudulent or legitimate). While the rest of the twelve features are anonymized via a principal component analysis (PCA) transform. For ethical and legal concerns, the PCA approach is used to conceal the cardholders' private information.

*Table 1. Dataset sample*

| DOMAIN | STATE | ZIP CODE | Time 1 | Time2 | VIS1 | VIS2 | XRN 1 | XRN 2 | XRN 3 | XRN 4 | XRN 5 | VAR 1 | VAR 2 | VAR3 | VAR 4 | VAR 5 | TRN_ AMT | TRN_ AMT | TRN_ TYPE |
|---|---|---|---|---|---|---|---|---|---|---|---|---|---|---|---|---|---|---|---|
| | | | | | | | | | | | | | | | | | | | |

| | | | | | | | | | | | | | | | | | | |
|---|---|---|---|---|---|---|---|---|---|---|---|---|---|---|---|---|---|---|
| CDRZLKA JLJVQHC N.COM | AO | 675 | 12 | 12 | 1 | 0 | 0 | 1 | 1 | 0 | 1 | 2 | 1 | 16.680 | 34 | 0 | 12.95 | 12.95 | LEGIT |
| NEKSXUK .NET | KK | 680 | 18 | 18 | 1 | 0 | 0 | 0 | 0 | 0 | 1 | 3 | 0 | 37.880 | 23 | 0 | 38.85 | 38.85 | LEGIT |
| XOSOP.C OM | UO | 432 | 3 | 3 | 1 | 0 | 0 | 1 | 1 | 0 | 1 | 3 | 1 | -9.080 | 19 | 2 | 38.85 | 38.85 | LEGIT |
| TMA.COM | KR | 119 | 23 | 23 | 0 | 0 | 1 | 0 | 0 | 0 | 3 | 0 | 0 | -6.392 | 18 | 0 | 11.01 | 11.01 | LEGIT |
| VUHZRNB .COM | PO | 614 | 9 | 9 | 0 | 0 | 0 | 1 | 0 | 0 | 1 | 3 | 0 | 42.512 | 7 | 0 | 12.95 | 12.95 | LEGIT |

A. Feature selection

The dataset contains non-numeric values. As shown in the Table 1, there are two columns for time and two columns for the amount of the transaction. To investigate whether these columns have the same values or not we perform the heat map for them in Figure 1.

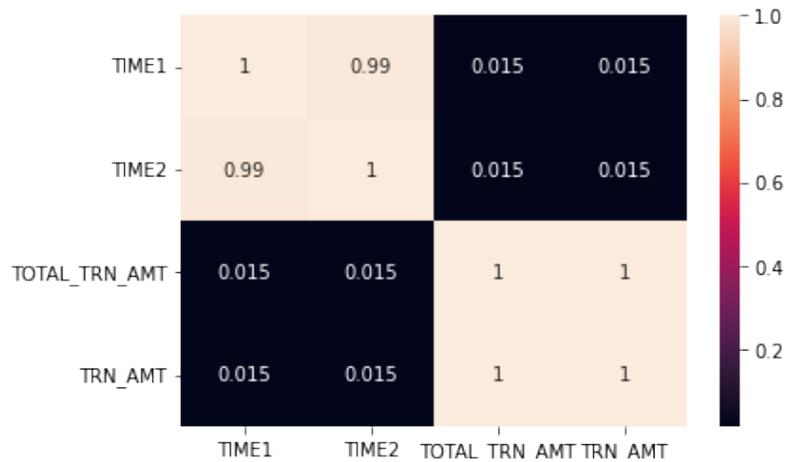

*Figure 1. Heat map of four features to be further investigated.*

Figure 1 shows that the Time1 and Time 2 features are quite the same (i.e., correlation of about 0.99) and this is logical because these features belong to the start and finish time for the transaction. The heat map also shows that the TOTAL_TRN_AMT, and TRN_AMT features are the same (i.e., correlation of about 1). So at this phase, we decided the following according to previous studies [29]–[31]:

- Delete TOTAL_TRN_AMT feature.
- Delete Domain, and State features as they lack to provide logical correlations with classification.
- Convert the TRN_TYPE feature, which is the target classes, into zero (not fraud) and 1 (fraud), to be used as input for binary classifier.

B. Data preprocessing

The original dataset contains details of total 94,682 transactions; of which about 87,562 transactions were not fraud with a percentage of (97.7%), and about 2,052 were fraud ones with a percentage of (2.3%). Data from credit card transactions are primarily distinguished by an unusual phenomenon. Transactions that are both legitimate and fraudulent frequently have the same characteristics. Fraudsters continually find new ways to imitate the spending habits of legitimate cards. The characteristics of legitimate and fraudulent behavior are thus continually changing. Because of the imbalanced dataset would produce various unexpected behaviors; it is not appropriate to use it directly as the input of our model. As shown in Figure 2, there is a huge contrast in numbers between positive and negative samples for fraud and legitimate classes, it is very likely that positive samples (fraudulent class) may be mistaken and the classifier will be biased to predict the negative (legitimate) class only [32].

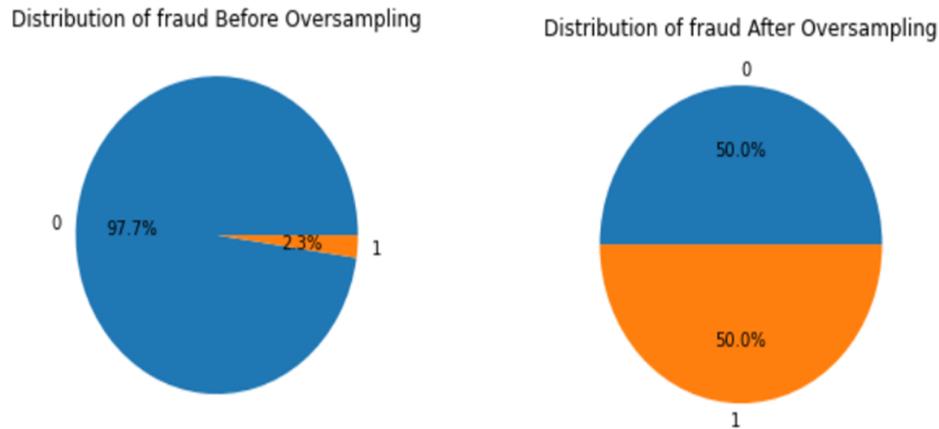

Figure 2. The distribution of transaction type (not fraud is 0, and fraud is 1)

The solution, in this case, is to perform over-sampling or under-sampling techniques. The synthetic minority over-sampling technique (SMOTE) is the most used one in the literature [33]–[36]. It includes creating synthetic samples and oversampling the minority class. According to the required number of synthetic samples, neighbors from the k-nearest neighbors would first be randomly selected, and then one sample would be generated in the general direction of each of the selected neighbors. then determine how much the feature vector (sample) under examination differs from its neighbor, multiply the difference by a random number between 0 and 1, then add it to the feature vector under consideration [37]–[39]. After performing the SMOTE over-sampling technique, the data distribution for both classes was a 50:50 ratio with a count of about 87,562 transactions for each class. Finally, we check for duplicated rows, and about 5,068 were found. we deleted them and get the dataset ready to perform the random forest classifier.

IV. Methodology

Random forest is a machine-learning approach for supervised learning based on ensemble learning. Ensemble learning is an algorithm that generates predictions through several iterations of the same model or a comparable model. Similar to this, the random forest algorithm employs numerous algorithms, or different decision trees, creating a forest of trees as a result. Both regression and classification tasks can be completed using the random forest technique [40]. Compared to the decision tree, the random forest has an advantage because it breaks the bad tendency of overfitting the training set. Each node of the decision tree is split on a chosen feature from a randomly selected subset of the entire feature set once a decision tree has been formed and a subset of the training set has been randomly sampled to train each individual tree. Because each tree in a random forest is trained independently of the others, training is incredibly quick even for big data sets with numerous characteristics and data occurrences. It has been discovered that the Random Forest algorithm is resistant to overfitting and offers a good approximation of the generalization error [41], [42]. The random forest algorithm was chosen because it acts as a binary classifier, which makes it suitable for credit card fraud detection, as soon as any transaction is classified as one of two classes (0 or 1) fraud or not a fraud. The methodology adopted in this study is illustrated in Figure 3. As shown in Figure 3, the first step is to prepare the dataset for the random forest implementation, after that enhancement of the algorithm will be performed to gain the highest possible accuracy. Finally, evaluate the result using different metrics which will be discussed in the result section.

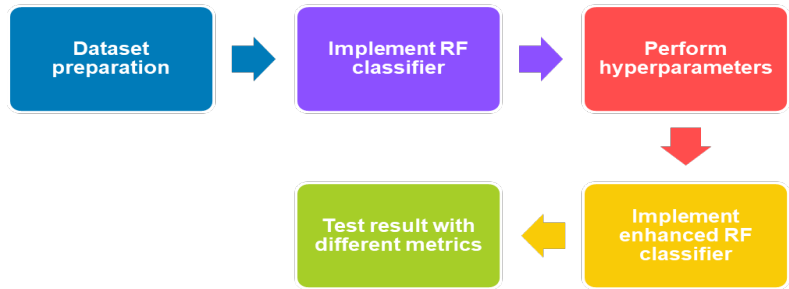

*Figure 3. Framework for predicting fraud in credit card transactions.*

V. Results and discussions

A. Experiment

The experimental work was conducted using Anaconda Jupyter V3, and the code was written using python language. The aim of the experiment was to offer a reliable fraud detection classifier that can successfully classify and detect fraud transactions. The dataset was split into training and testing data. performing SMOTE oversampling technique to avoid bias when running the random forest classifier and overcome the problem of the imbalanced dataset. Finally, we run the grid search to find the best parameters to run the classifier again to gain the highest possible accuracy.

B. Metrics and results

Although dealing with a highly skewed dataset, testing the algorithm showing only the accuracy is not sufficient to show the reliability of the algorithm. As a result, the confusion matrix, precision, recall, F1 score, and Receiver Operator Characteristic (ROC) curve measures were applied.

1) Confusion matrix is a metric that provides details on classes that were successfully and erroneously classified. The output of the confusion matrix is a 2X2 matrix that illustrates the value of TP, TN, FP, and FN. True Positive (TP), illustrating that your model predicted positive and it's true. True Negative (TN) means that you predicted negative and it's true. False Positive (FP) (Type 1 Error) means you predicted positive and it's false. False Negative (FN) (Type 2 Error) means you predicted negative and it's false [43]. The confusion matrix of our classifier is shown in Figure 4. The matrix is showing the number of correct predictions (True Positives TP=83736 and True Negatives TN=87242) and the number of incorrect predictions (False Positives FP=3826 and False Negatives FN=320) made by the classifier.

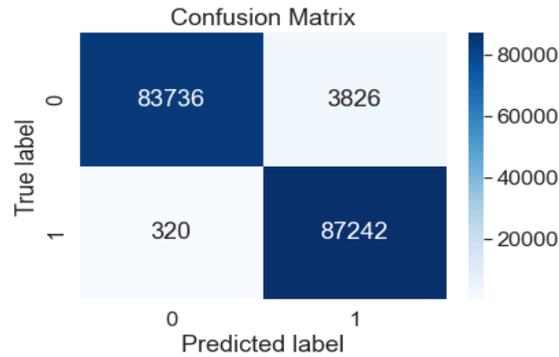

*Figure 4. Resulted confusion matrix.*

2) Accuracy, precision, recall, and F1 score measures. Accuracy is the proportion of correctly anticipated results. The overall accuracy of the classifier can be calculated by adding the number of true positives and true negatives and dividing by the total number of predictions. In that sense, the classifier is performing well in terms of accuracy. However, accuracy is not always the best metric to evaluate a classifier's performance, especially when the classes are imbalanced. The classifier's performance should be evaluated in terms of other metrics including precision, recall, and F1-score to get a better understanding of the classifier's performance. Precision is measured by the number of correctly identified outputs (Precision = [TP/ (TP+FP)]). Recall is the percentage of True Positives that the model properly identified (Recall = [TP/(TP+FN)]). While the harmonic mean of precision and recall is called F1-score (F1 score = [ 2*precision*recall / (precision + recall)]). In summary, the classifier has high accuracy, precision, recall, and F1-score, which are good results.

*Table 2. Classification report*

| Class | Precision | Recall | F1-score |
|---|---|---|---|
| **0 (Not fraud)** | 1.00 | 0.96 | 0.98 |
| **1 (Fraud)** | 0.96 | 1.00 | 0.98 |

As shown in Table 2, the classifier gained an accuracy of 98% to predict the fraud transaction of all transactions, with precision, recall, and F1 score of 100%, 96%, and 98% respectively for class 0 which indicates non-fraud transactions. And 96%, 100%, and 98% respectively for class 1 which indicates fraud transactions.

3) The ROC curve is a graphical plot that illustrates the diagnostic ability of a binary classifier system as its discrimination threshold. By plotting the true positive rate (TPR) also known as sensitivity, and recall against the false positive rate (FPR) also known as the probability of false alarms [44].

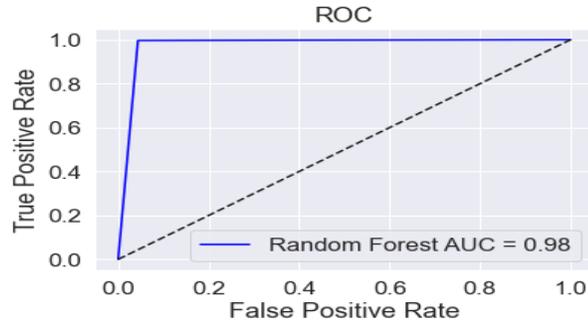

*Figure 5. Random forest ROC*

AUC has a value between 0 and 1, with a value of 1 indicating perfect performance and a value of 0.5 indicating a classifier no better than random guessing. As shown in Figure 5, the result of the random forest classifier AUC was 0.98. It means that the classifier is able to correctly classify 98% of the positive examples as positive and negative examples as negative. It also means that the false positive rate is low, which is the rate at which the classifier incorrectly classifies negative examples as positive. Our results are in the same line as other studies [19], [21], [22], [32], [43], [45]–[47]. Nonetheless, we believe our model is relatively easier to apply and can overcome the issue of imbalanced data in fraud detection.

VI. Conclusion

This study implemented an enhanced random forest algorithm using hyperparameters to find the best values to run the classifier and gains the highest possible accuracy. We performed the SMOTE oversampling technique to overcome the imbalanced dataset issue. The result showed that the classifier was able to predict fraudulent transactions with an accuracy of 98%. We evaluate our results using different metrics such as recall, precision, F1 score, and the ROC curve to improve that our classifier is not biased because of the imbalanced dataset and that it's able to detect both classes (fraud, not fraud) with a high accuracy level. Results showed that our model, due to its relatively straightforward application, is well-suited to addressing the challenge of imbalanced data in fraud detection.